\documentclass{article}
\usepackage{spconf,amsmath,graphicx}

\usepackage{amsmath}
\usepackage{amsmath,bm}
\usepackage{amssymb}
\usepackage{multicol}
\usepackage{graphicx}
\usepackage{epstopdf}
\usepackage{booktabs}
\usepackage{cite}
\usepackage{color}
\usepackage{algorithm}
\usepackage{algorithmic}
\usepackage{dsfont}
\usepackage[Symbol]{upgreek}
\usepackage{psfrag}
\usepackage{setspace}
\usepackage{acronym}
\usepackage{pdfpages}
\usepackage{subfigure}
\usepackage{stfloats}
\usepackage{url}
\usepackage{bm}
\usepackage{multirow}

\DeclareMathAlphabet{\mathsfbr}{OT1}{cmss}{m}{n}
\SetMathAlphabet{\mathsfbr}{bold}{OT1}{cmss}{bx}{n}
\DeclareRobustCommand{\msf}[1]{%
  \ifcat\noexpand#1\relax\msfgreek{#1}\else\mathsfbr{#1}\fi
}

\makeatletter
\newcommand{\msfgreek}[1]{\csname s\expandafter\@gobble\string#1\endcsname}
\makeatother

\DeclareFontEncoding{LGR}{}{} 
\DeclareSymbolFont{sfgreek}{LGR}{cmss}{m}{n}
\SetSymbolFont{sfgreek}{bold}{LGR}{cmss}{bx}{n}
\DeclareMathSymbol{\salpha}{\mathord}{sfgreek}{`a}
\DeclareMathSymbol{\sbeta}{\mathord}{sfgreek}{`b}
\DeclareMathSymbol{\sgamma}{\mathord}{sfgreek}{`g}
\DeclareMathSymbol{\sdelta}{\mathord}{sfgreek}{`d}
\DeclareMathSymbol{\sepsilon}{\mathord}{sfgreek}{`e}
\DeclareMathSymbol{\szeta}{\mathord}{sfgreek}{`z}
\DeclareMathSymbol{\seta}{\mathord}{sfgreek}{`h}
\DeclareMathSymbol{\stheta}{\mathord}{sfgreek}{`j}
\DeclareMathSymbol{\siota}{\mathord}{sfgreek}{`i}
\DeclareMathSymbol{\skappa}{\mathord}{sfgreek}{`k}
\DeclareMathSymbol{\slambda}{\mathord}{sfgreek}{`l}
\DeclareMathSymbol{\smu}{\mathord}{sfgreek}{`m}
\DeclareMathSymbol{\snu}{\mathord}{sfgreek}{`n}
\DeclareMathSymbol{\sxi}{\mathord}{sfgreek}{`x}
\DeclareMathSymbol{\somicron}{\mathord}{sfgreek}{`o}
\DeclareMathSymbol{\spi}{\mathord}{sfgreek}{`p}
\DeclareMathSymbol{\srho}{\mathord}{sfgreek}{`r}
\DeclareMathSymbol{\ssigma}{\mathord}{sfgreek}{`s}
\DeclareMathSymbol{\stau}{\mathord}{sfgreek}{`t}
\DeclareMathSymbol{\supsilon}{\mathord}{sfgreek}{`u}
\DeclareMathSymbol{\sphi}{\mathord}{sfgreek}{`f}
\DeclareMathSymbol{\schi}{\mathord}{sfgreek}{`q}
\DeclareMathSymbol{\spsi}{\mathord}{sfgreek}{`y}
\DeclareMathSymbol{\somega}{\mathord}{sfgreek}{`w}

\DeclareMathSymbol{\svarsigma}{\mathord}{sfgreek}{`c}

\DeclareMathSymbol{\sGamma}{\mathalpha}{sfgreek}{`G}
\DeclareMathSymbol{\sDelta}{\mathalpha}{sfgreek}{`D}
\DeclareMathSymbol{\sTheta}{\mathalpha}{sfgreek}{`J}
\DeclareMathSymbol{\sLambda}{\mathalpha}{sfgreek}{`L}
\DeclareMathSymbol{\sXi}{\mathalpha}{sfgreek}{`X}
\DeclareMathSymbol{\sPi}{\mathalpha}{sfgreek}{`P}
\DeclareMathSymbol{\sSigma}{\mathalpha}{sfgreek}{`S}
\DeclareMathSymbol{\sUpsilon}{\mathalpha}{sfgreek}{`U}
\DeclareMathSymbol{\sPhi}{\mathalpha}{sfgreek}{`F}
\DeclareMathSymbol{\sPsi}{\mathalpha}{sfgreek}{`Y}
\DeclareMathSymbol{\sOmega}{\mathalpha}{sfgreek}{`W}

\DeclareRobustCommand{\mcal}[1]{%
  \ifcat\noexpand#1\relax\mathnormal{#1}\else\cal{#1}\fi
}
\DeclareRobustCommand{\BM}[1]{%
  \ifcat\noexpand#1\relax\bm{\boldUppercaseItalicGreek{#1}}\else\bm{#1}\fi
}
\makeatletter
\newcommand{\boldUppercaseItalicGreek}[1]{\csname var\expandafter\@gobble\string#1\endcsname}
\makeatother
\newcommand{\rv}[1]{\MakeLowercase{\msf{#1}}}
\newcommand{\RV}[1]{\bm{\MakeLowercase{\msf{#1}}}}

\newcommand{\V}[1]{\bm{#1}}
\newcommand{\M}[1]{\BM{#1}}
\newcommand{\Set}[1]{\mcal{#1}}

\newtheorem{remark}{Remark}
\title{A DRL BASED DISTRIBUTED FORMATION CONTROL SCHEME WITH STREAM-BASED COLLISION AVOIDANCE}
\name{Xinyou Qiu, Xiaoxiang Li, Jian Wang, Yu Wang, Yuan Shen }
\address{Beijing National Research Center for Information Science and Technology\\
Department of Electronic Engineering, Tsinghua University, Beijing 100084, China\\[0.1em]
Email: \{qxy18, lxx17\}@mails.tsinghua.edu.cn, \{jian-wang, yu-wang, shenyuan\_ee\}@tsinghua.edu.cn}
\vspace{-0.5cm}

\begin{document}
%
\maketitle
\begin{abstract}
Formation and collision avoidance abilities are essential for multi-agent systems. Conventional methods usually require a central controller and global information to achieve collaboration, which is impractical in an unknown environment. In this paper, we propose a deep reinforcement learning (DRL) based distributed formation control scheme for autonomous vehicles. A modified stream-based obstacle avoidance method is applied to smoothen the optimal trajectory, and onboard sensors such as Lidar and antenna arrays are used to obtain local relative distance and angle information. The proposed scheme obtains a scalable distributed control policy which jointly optimizes formation tracking error and average collision rate with local observations. Simulation results demonstrate that our method outperforms two other state-of-the-art algorithms on maintaining formation and collision avoidance.
\end{abstract}
\vspace{-2mm}
\section{INTRODUCTION}
\label{sec:intro}
\vspace{-2mm}
With the maturity and promotion of the Internet of Things, collaboration among agents becomes more and more imperative\cite{XioWuSheWin:C19,WinConMazSheGifDarChi:J11,AmiAsiMoh:J18}. By deploying a multiple-agent system, sophisticated tasks such as harsh area exploration, disaster rescuing and
map reconstruction become more tractable since each agent can be more concentrated on its own subtask. Among the
key technologies of the multi-agent collaboration, formation control is the most important and practical one, through which
we can assign each agent's relative position, (i.e. a desired formation ``shape''\cite{LiuWanWanShe:J20}), and enable the multiple agents to
explore the environment more efficiently. Furthermore, the robustness can be enhanced significantly by applying proper
distributed operation and node failure tolerance methods.

In terms of formation control, collision avoidance is a critical task since agents often have to simultaneously maintain the
desire formation and prevent collisions. Artificial potential field (APF) is a common solution for collision avoidance problems\cite{SamBavRamPueCam:C18,LiWu:J20,OkaAke:J15}. By constructing a potential field function, the agent can calculate the interaction forces with the obstacles and the destination. However, APF suffers from the local minima problem in complex environments with multiple obstacles, which could result in non-smooth motion or even be trapped in the saddle points\cite{OkaAke:J15}. Some researchers have focused on applying the concepts of fluid mechanics, such as stream function
methods, to generate smoother trajectories in exclusion of local minimum\cite{WayMur:C03,WanCheFanMa:J14}. But the global position information of the agent
and obstacles is required by these methods, which may be impractical in harsh cases. Deep reinforcement learning (DRL)\cite{KhaHerLewPipMel:J14,SutBarWil:J92,WenCheFenZho:J14} offers a promising solution to such a distributed formation control problem in arbitrary unknown systems due to its optimal-adaptive and model-free properties. By designing a proper cost function, agents can iteratively optimize
policy by continuously exploring the environment even with local observation.

In this paper, we put forward a DRL scheme to train a formation control system with stream-based collision avoidance. The main contributions of our work are as follows:

1) We put forward a distributed formation control scheme with a modified stream-based collision avoidance policy. The
proposed scheme requires only local observation for each agent and is easily achievable for real systems.

2) We design a DRL model, based on deep deterministic policy gradient (DDPG), to train the distributed policy in continuous action-state space. The decision-making and control layers are tightly coupled in the proposed model and thus guarantee a better collaboration
between multiple agents. Simulation results show good performance compared with the existing obstacle-avoiding algorithms.

\emph{Notations}: Throughout this paper, variables, vectors, and matrices are written as italic letters $x$, bold italic letters $\V{x}$, and bold capital italic letters $\M{X}$, respectively. Random variables and random vectors are written as sans serif letter $\rv{x}$ and bold letters $\RV{x}$, respectively. The notation $\mathbb{E}_{\RV{x}} \{\cdot \}$ is the expectation operator with respect to the random vector $\RV{x}$, and $\mathds{1}(\cdot)$ is the indicator function which equals $1$ if the condition is true and equals $0$ otherwise.
\vspace{-4mm}
\section{PRELIMINARIES}
\label{sec:preliminaries}
\vspace{-2mm}
In fluid dynamics, the stream function is often introduced to analyze a particle's behavior in a 2-D incompressible flow field. For example, streamlines plotted from different stream functions represent the trajectories of particles inside the field. To obtain the stream function for an incompressible flow in a 2-D $x-y$ plane, we first consider the continuity equations given by
\vspace{-2mm}
\begin{equation}
\begin{aligned}\label{continuous equation}
\frac{\partial u}{\partial x}+\frac{\partial v}{\partial y}=0,
\end{aligned}
\vspace{-1mm}
\end{equation}
where $u$, $v$ are the fluid velocity in $x$, $y$ axis direction, respectively. The velocity field in a 2-D incompressible flows always satisfy (\ref{continuous equation}). Define the function $\psi(x,y)$ as
\vspace{-2mm}
\begin{equation}
\begin{aligned}\label{stream function}
u=\frac{\partial \psi}{\partial y},v=-\frac{\partial \psi}{\partial x},
\end{aligned}
\vspace{-1mm}
\end{equation}
then the continuity equation becomes
\vspace{-2mm}
\begin{equation}
\begin{aligned}\label{continuous equation 2}
\frac{\partial }{\partial x}\frac{\partial \psi}{\partial y}+\frac{\partial }{\partial y}(-\frac{\partial \psi}{\partial x})=\frac{\partial ^2 \psi}{\partial x\partial y}-\frac{\partial ^2 \psi}{\partial y\partial x}=0.
\vspace{-1mm}
\end{aligned}
\end{equation}
Note that if there exists a function $\psi$ which fulfills (\ref{continuous equation 2}), it can be used to obtain streamlines $\psi=C$ for every point in the flow field. The points on the same streamlines share the same constant value $C$.

For a flow field with a cylinder-shaped obstacle locating at the origin, we can obtain the stream function by treating it as a combination of the uniform flow and the doublet flow\cite{WanCheFanMa:J14}. Assume the strength of flow is $U$, the compound stream function is given as
\vspace{-2mm}
\begin{equation}
\begin{aligned}\label{compound stream function}
\psi = \psi_{\text{uniform flow}} + \psi_{\text{doublet}} = Uy-U(\frac{r^2y}{x^2+y^2}).
\end{aligned}
\vspace{-1mm}
\end{equation}
By following the assigned streamline, the agent can avoid collision smoothly since no local minima exists in the field.
\vspace{-4mm}
\section{SYSTEM MODEL}
\label{sec:pagestyle}
\begin{figure}
  \centering
  \includegraphics*[scale=0.5]{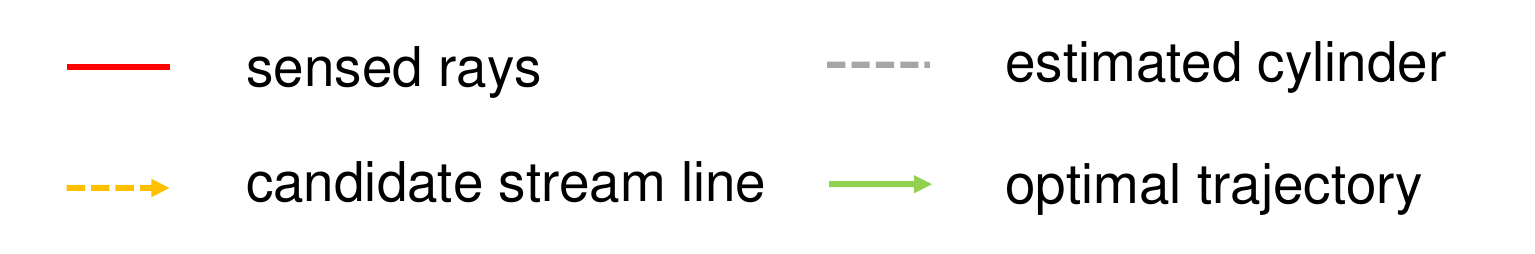}\\
  \hspace{-4mm}
  \subfigure[system model and geometric relationship]{
  \includegraphics*[scale=0.35]{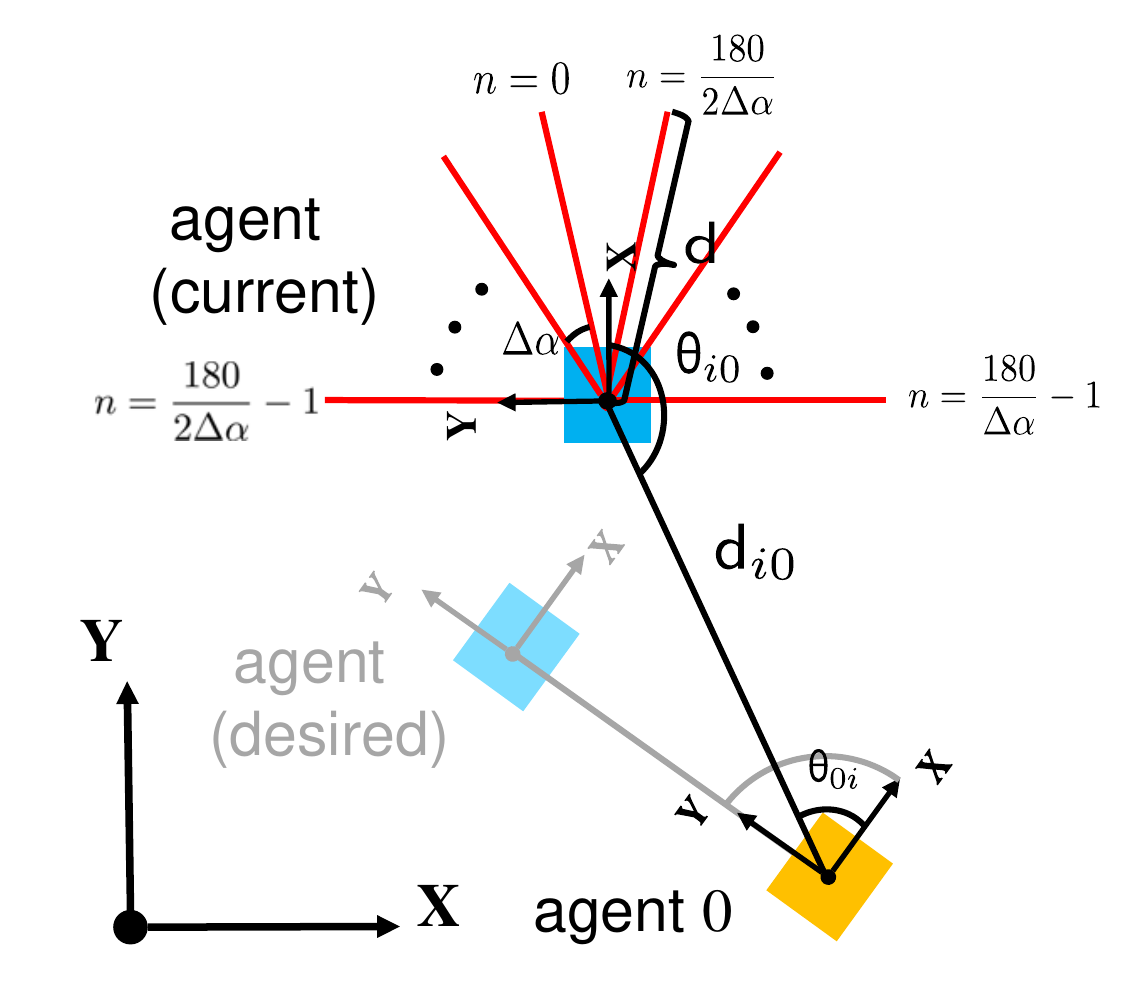}
  \label{fig:system model}
  }
  \subfigure[proposed stream-based avoiding policy]{
  \includegraphics*[scale=0.35]{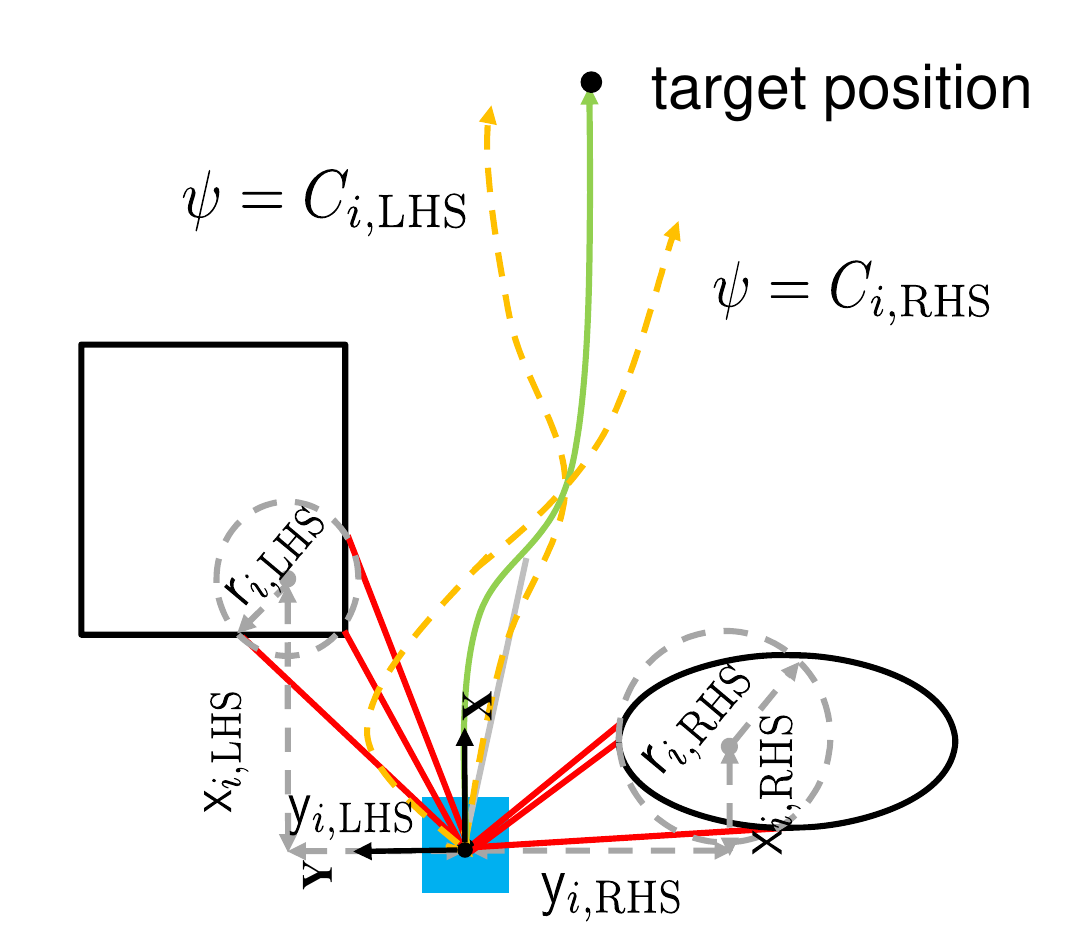}
  \label{fig:avoiding model}
  }
  \vspace{-4mm}
  \caption{ The schematic diagram of our system and method.}
  \vspace{-4mm}
\end{figure}

\vspace{-2mm}
Consider a multi-agent navigation system with $1$ virtual navigator agent and $N-1$ follower agents moving in a 2-D Euclidean plane at time $k\in[0,T_{\text{max}})$ with the following dynamic model\\
\vspace{-2mm}
\begin{equation}\label{system model}
\begin{aligned}
\RV{x}_i(k+1)=&[x_i(k)+\Delta x_i(k),y_i(k)+\Delta y_i(k), v_i(k)+ka_i(k),\\
&\alpha_i(k)+k\omega_i(k), \omega_i(k) + k\beta_i(k)]^\intercal+\RV{w}_i(k)\\
\triangleq &f(\RV{x}_i(k),\V{u}_i(k))+\RV{w}_i(k)
\end{aligned}
\vspace{-1mm}
\end{equation}
where $i=0,...,N-1$. $i=0$ is the virtual navigator agent and the rest are follower agents. $\V{p}_i(k)=[x_i(k),y_i(k)]^\intercal\in\mathcal{R}^2$ and $\alpha_i\in(-\pi,\pi]$ are the position and the orientation of agent $i$ in the global coordinate system, $v_i(k)$ and $\omega_i(k)$ are the velocity and the angular velocity. $\Delta x_i(k)=\frac{v_i(k)}{\omega_i(k)}\left[\text{sin}(\alpha_i(k)+k\omega_i(k))-\text{sin}(\alpha_i(k))\right]$, $\Delta y_i(k)=\frac{v_i(k)}{\omega_i(k)}\\
\left[\text{cos}(\alpha_i(k)+k\omega_i(k))-\text{cos}(\alpha_i(k))\right]$, $\V{u}_i(k)=[a_i(k),\beta_i(k)]^\intercal\\
\in\mathcal{R}^2$ are the control inputs of the acceleration and angular acceleration, and $\RV{w}_i(k)\sim\mathcal{N}(0,\M{C})$ are independent white Gaussian state noises with zero means and covariance matrix $\M{C}$ in the local coordinate system. System (\ref{system model}) is assumed stabilizable, i.e. there exists a continuous control $\V{u}_i$ such that the system is asymptotically stable. The communication graph $\Set{G}=(\Set{V},\Set{E})$ is assumed as an undirected connected graph if the adjacent agents stay in the connection zone $\delta_i$. The set of neighbors of $i\in \mathcal{V}$ is defined as $\mathcal{N}_i:=\{j\in\mathcal{V}:(i,j)\in\mathcal{E}\}$. For any agent $i$, the relative position set is denoted as $\mathcal{P}_i=\{[d_{ij}, \theta_{ij}]^\intercal, \forall j \in \mathcal{N}_i\}$, which includes the distance and angle to all adjacent agents that can be communicated through antenna array-based sensors. Agent $0$ will continuously broadcast $\theta_{0j}(k)$ to every connectable agent through communication. To detect the collision, lidar-based distance sensors are equipped on the follower agents. The sensors provide distance measurements $\RV{d}=\V{d}+\RV{n}$ with an angle resolution of $\Delta\alpha$, where $d_\text{min}\leq d \leq d_\text{max} \forall d \in \V{d}$ and $\RV{n}\sim\mathcal{N}(0,\M{C}_{\text{lidar}})$ denotes additive white Gaussian noises with zero means and covariance matrix $\M{C}_{\text{lidar}}$, as Fig. \ref{fig:system model} illustrates.

Agent $0$ is responsible for providing a trajectory which navigates the swarm toward the goal. The follower agents must maintain their predefined relative position $\V{\eta}_i=[d_i\text{cos}(\theta_i),d_i\text{sin}(\theta_i)]^\intercal$ with respect to agent $0$ while avoiding collision. Define the tracking error for agent $i$ as
\vspace{-2mm}
\begin{equation}
\begin{aligned}\label{tracking error 1}
\V{e}_i(k)=\V{p}_i(k)-\V{p}_{0}(k)-\V{\eta}_i.
\end{aligned}
\vspace{-1mm}
\end{equation}
Note that $d_{i0}(k)$ and $\theta_{i0}(k)$ are available through agent $0$'s broadcast, (\ref{tracking error 1}) can be rewritten as following form
\vspace{-2mm}
\begin{equation}
\begin{aligned}\label{tracking error 2}
\V{e}_i(k)=\V{z}_{i}(k)-\V{\eta}_i
\end{aligned}
\vspace{-1mm}
\end{equation}
where
\vspace{-2mm}
\begin{equation}
\begin{aligned}\label{tracking error 3}
\V{z}_{i}(k)&=\V{p}_i(k)-\V{p}_{0}(k)\\
&=[d_{0i}(k)\text{cos}(\theta_{0i}(k)),d_{0i}(k)\text{sin}(\theta_{0i}(k))]^\intercal
\end{aligned}
\vspace{-1mm}
\end{equation}
is the relative displacement vector between follower agent $i$ and agent $0$. Although $\V{p}_{0}(k),k=0,...,T_{\text{max}}$ is unknown for agent $i$, tracking the trajectory is viable through the observed relative position. To minimize the tracking error, we introduce the following cost function\cite{LiuMenPenLew:J20}
\vspace{-2mm}
\begin{equation}
\begin{aligned}\label{formation cost}
r_{i, \text{tracking}}(k)=\V{e}^\intercal_i(k)\M{Q}_e\V{e}_i(k),
\end{aligned}
\vspace{-1mm}
\end{equation}
where $\M{Q}_e$ is a positive definite matrix.
\vspace{-2mm}
\section{APPROACH AND IMPLEMENT DETAILS}
\label{sec:approach}
\vspace{-2mm}
\subsection{Distributed stream-based collision avoidance}
\vspace{-2mm}
Traditionally, stream-based collision avoidance requires the agent to follow a set of virtual leader's trajectory on the desired streamline. This set will be set at the beginning according to the coordinates of the obstacle. However, most unmanned vehicles are designed to explore unknown areas where prior knowledge of the obstacles is unavailable. Therefore, we introduce a safe distance range $\V{d}_{\text{safe}}=[d_{\text{risk}},d_{\text{stop}}]^\intercal$, where $d_{\text{risk}}$ is the risk distance that the agent should start the avoiding behavior, and $d_{\text{stop}}$ is the minimum braking distance for an agent to stop from the highest speed\cite{ParHuh:J16}. To apply stream-based collision avoidance on our system, several assumptions are necessary:\\
1) The velocity of the agent is nonnegative, i.e., $v_i(k)\ge0$,\\
2) The minimum horizontal projection length of obstacles is at least  $2d_{\text{risk}}\text{sin}(\Delta\alpha)$, so that it can be detected by at least three rays.

When an obstacle is detected by agent $i$, the direction of detected rays form a discrete interval whose indices can be denoted as a set ${\Set{A}}_i=\{n_\text{start},n_\text{start}+1,...,n_\text{end}-1,n_\text{end}\}$, where $|\alpha_n-\alpha_{n-1}|=\Delta\alpha, \forall n>0$.
The endpoint of each ray $n$ is denoted as $\RV{p}_n(k)=[\rv{d}_{n}(k)\text{cos}(\alpha_{n}(k)),\rv{d}_{n}(k)\text{sin}(\alpha_{n}(k))]^\intercal, n\in[n_\text{start}, n_\text{end}]$, where $n_\text{start}, n_\text{end}$ are the indices of the start and end ray in the interval. We can always find a shortest ray $m=\arg\min  ||\RV{p}_{m}(k)||, m\in{\Set{A}}_i\backslash\{n_\text{start}, n_\text{end}\}$ in the interval, where $m$ represents the index of the shortest ray except the start and end ray. The three endpoints $\RV{p}_m(k)$, $\RV{p}_{n_\text{start}}(k)$ and $\RV{p}_{n_\text{end}}(k)$ can form a triangle as long as they are not on a line.

To deal with multiple obstacles, we divide the front semicircle into left-hand side (LHS) and right-hand side (RHS) to represent two flow fields, whose interval sets are denoted as ${\Set{A}}_{i,\text{LHS}}= \{n\in{\Set{A}}_i|\alpha_n>0\}$ and ${\Set{A}}_{i,\text{RHS}}= \{n\in{\Set{A}}_i|\alpha_n<0\}$. The two fields are independent and consider only the foremost interval sets (minimum $n_\text{start}$ on either side). By finding the circumcenter of the triangle, we can obtain a virtual cylinder with radius $\RV{r}_{i,\text{obstacle}}=[\rv{r}_{i,\text{LHS}},\rv{r}_{i,\text{RHS}}]^\intercal$ and relative position $\RV{p}_\text{i,cyl}(k)=[\RV{x}^\intercal_\text{i,cyl}(k),\RV{y}^\intercal_\text{i,cyl}(k)]^\intercal$ where $\RV{(\cdot)}_{i,\text{cyl}}(k)=[\rv{(\cdot)}_{i,\text{LHS}},\rv{(\cdot)}_{i,\text{RHS}}]^\intercal$ under the agent's coordinate system, as Fig. \ref{fig:avoiding model} illustrates.

Based on the assumptions, it is impossible for any obstacle to be ahead of the agent after being avoided, i.e. during avoidance, the virtual cylinder is pretended identical if $|(\alpha_{n_\text{start}}(k))|>|(\alpha_{n_\text{start}}(k-1))|$. Given that the cylinders being the same, $-\RV{p}_{i,\text{cyl}}(k)$ can represent the displacements of agent $i$ to the origin of virtual cylinders with radius $\RV{r}_{i,\text{obstacle}}(k)$. To this end, the agent is capable of calculating $\RV{c}_i=[\rv{c}_{i,{\text{LHS}}}(k),\rv{c}_{i,{\text{RHS}}}(k)]^\intercal$ in (\ref{compound stream function}) with local observation.

The desired stream values of the two flow fields are denoted as $\V{c}_{i,\text{desired}}=[C_{i,\text{LHS}},C_{i,\text{RHS}}]^\intercal$. To ensure the availability of the streamlines for different agent sizes, $\V{c}_{i,\text{bound}}$ is given based on $[x_{i,\text{LHS}},y_{i,\text{LHS}},x_{i,\text{RHS}},y_{i,\text{RHS}}]^\intercal=[0,-d_{\text{stop}},0,d_{\text{stop}}]^\intercal$. From \cite{WayMur:C03}, we have observed that following the streamline is in fact to find $\RV{c}_i=\min||\V{c}-\V{c}_{i,\text{desired}}||$. On top of that, our algorithm tracks $\V{c}_{i,\text{desired}}$ directly instead of predefined point sets of the streamline. The avoidance cost function can be formulated as
\vspace{-2mm}
\begin{equation}
\begin{aligned}\label{avoiding cost}
r_{i,\text{avoiding}}(k)=\sum^{\text{len}(\V{c}_i)-1}_{s=0} \mathds{1}(avoid_{s}(k))(\RV{C}_{i}(s)-\\\V{c}_{i,\text{desired}}(s))^2\left(\frac{1}{||\RV{p}_m(k)||}-\frac{1}{d_{\text{risk}}}\right),
\end{aligned}
\vspace{-1mm}
\end{equation}
where $avoid_{s}(k)$ is the avoidance flag, $(\RV{c}_{i}(s)-\V{c}_{i,\text{desired}}(s))^2$ calculates the stream value error. The cost is multiplied with a potential field-based cost \cite{SamBavRamPueCam:C18} in case $\RV{c}_{i}(s)$ changes too subtly when $\RV{y}_{i,\text{cyl}}(s)\approx0$. (\ref{avoiding cost}) reliefs the strong penalty when the agent tries to go through a faster path but with more obstacles. Thus, the agent can recover the formation faster while remaining safe in a multi-obstacle environment. The entire policy is shown in Algorithm \ref{avoiding behavior}.
\vspace{-2mm}
\begin{algorithm}[h]
 \caption{Avoidance Decision Policy}
 \label{avoiding behavior}
 \begin{algorithmic}[1]
    \STATE obtain distance measurements of agent $i$
    \FOR {$s$ in $0,..., \text{len}(\V{c}_i)-1$}
        \STATE initialize $avoid_{s}(k)$ as false
        \STATE set $avoid_{s}(k)$ true and calculate $\RV{p}_{i,\text{cyl}}(s)$, $\RV{r}_{i,\text{obstacle}}(s)$ if detected
        \IF{$avoid_{s}(k)$}
            \IF{not $avoid_{s}(k-1)$}
                \STATE calculate $\V{c}_{i,\text{desired}}(s)$ by (\ref{compound stream function}), check availability with $\V{c}_{{s},\text{bound}}$
            \ELSE
                \STATE calculate $\RV{c}_{i}(s)$
                \STATE update $\V{c}_{i,\text{desired}}(s)\gets\RV{c}_{i}(s)$ if new obstacle detected
            \ENDIF
        \ENDIF
    \ENDFOR
    \RETURN{$r_{i,\text{avoiding}}(k)$ in (\ref{avoiding cost})}
 \end{algorithmic}
\end{algorithm}
\vspace{-6mm}
\begin{remark}
To apply the algorithm, default values should be set if no triangle can be formed by the detected results. Also the agent should choose a side to avoid once the obstacle overlaps LHS and RHS.
\vspace{-4mm}
\end{remark}
\subsection{DRL-based control policy optimization}
\vspace{-2mm}
We apply DDPG\cite{LilHunPriHeeEreTasSilWie:C03} to optimize the control policy without knowing the system model. A critic network $Q^{\V{\varpi}}(\RV{o},\V{u})$ and an actor network $\V{u}^{\V{\phi}}(\RV{o})$, parameterized by $\V{\varpi},\V{\phi}$, are created to evaluate the value of the current state and produce the control inputs.

The critic network of the entire system $Q(k)$ is defined as
\vspace{-2mm}
\begin{equation}
\begin{aligned}\label{cost sum}
Q(k)=\sum^{N-1}_{i=1}Q^{\V{\varpi}}_{i}(\RV{o}_i(k),\V{u}^{\V{\phi}}_i(k))=\sum^{N-1}_{i=1}\sum^{T_{max}}_{t=k}\gamma^{t-k}r_{i}(\RV{o}_i(k)),
\end{aligned}
\vspace{-1mm}
\end{equation}
where $\gamma$ is the discount factor of expected future cost, $r_{i}(\RV{o}_i(k))=r_{i,\text{tracking}}(k)+r_{i,\text{avoiding}}(k)$ is the sum of (\ref{formation cost}) and (\ref{avoiding cost}), $\RV{o}_i(k)=[\V{e}_i(k),\alpha_i(k),v_i(k),\omega_i(k),\mathcal{P}_i,\mathcal{P}_{\text{sensor}},\RV{o}_{\text{cyl}}]^\intercal$ is the observation vector, ${\Set{P}}_{\text{sensor}}=\{[\rv{d}_{n}(k),\alpha_{n}(k)]^\intercal, \forall n=\{{n_{\text{start}},m,n_{\text{end}}}\} \in \{{\Set{A}}_{i,\text{LHS}},{\Set{A}}_{i,\text{RHS}}\} \}$ includes the chosen detected results from the two sides. To promote a faster convergence, $\RV{o}_{\text{cyl}}(k)=[\RV{p}_{\text{cyl}}(k),\RV{r}_{\text{obstacle}}(k)]^\intercal$ is preprocessed from ${\Set{P}}_{\text{sensor}}$ instead of learning it by DDPG.
Since the formation system is homogeneous\cite{WenCheFenZho:J14} and we deploy the same policy on all follower agents, the policy can be deployed distributedly. The minimization of (\ref{cost sum}) and the policy is obtained by updating the network with \cite{LilHunPriHeeEreTasSilWie:C03}.
\vspace{-4mm}
\section{Simulation Results}
\vspace{-2mm}
\begin{table}
\renewcommand{\arraystretch}{1.3}
\caption{Simulation settings}
\vspace{-4mm}
\label{tab:setting}
\begin{center}
\scalebox{0.7}{
\begin{tabular}{|c|c|c|}
\hline
&name&value\\
\hline
\multirow{8}*{environment settings}&agent shape(size)& square($0.1$m)\\
&obstacle size& $r_{\text{obstacle}}\sim\mathcal{U}(0.1,0.5)$m\\
&lidar resolution& $\Delta\alpha=3^\circ$\\
&detect range, noise& $\V{d}\in[0, 2\text{m}],\RV{n}\sim\mathcal{N}(0,(0.2\text{m})^2\M{I})$\\
&maximum $v_i(v_0)$ & $0.5\text{m/s}(0.35\text{m/s})$\\
&maximum $|\omega_i|(|\omega_0|)$ & $|0.2\text{rad/s}|(|0.06\text{rad/s}|)$\\
& connection zone& $\delta_i=7$m\\
&safe distance $\V{d}_{\text{safe}}$& $[d_{\text{risk}}=0.7\text{m},d_{\text{stop}}=0.4\text{m}]^\intercal$\\
\hline
\multirow{7}*{hyperparameters}&network architecture& $(64,128,128)$ FC layers\\
&critic (actor) $lr$& $10^{-3}(10^{-4})$\\
&action space& $u^\phi=[u_0,u_1,u_2]^\intercal$ with softmax layers\\
&batch size& $1024$\\
&training episodes& $30000$\\
&episode steps& $250$\\
&step time& $0.1$sec\\
\hline
\end{tabular}}
\end{center}
\vspace{-8mm}
\end{table}
The proposed method is evaluated in an obstructive area. At most $5$ round obstacles are randomly scattered in a $14\times14\text{m}^2$ area around agent $0$, who provides the trajectory with navigation method in \cite{WanWanSheZha:J16}. $4$ follower agents are deployed around agent $0$ to form a circle with radius $2.1$m. The state transfer noise is $\RV{w}\sim\mathcal{N}(0,\M{C})$, where $\M{C}=\text{diag}\{10^{-2}\text{m}^2,10^{-2}\text{m}^2,10^{-4}(\text{m/s})^2,3.2\times10^{-4}\text{rad}^2,3.2\times10^{-6}(\text{rad/s})^2\}$. The rest environment settings and the hyperparameters of DDPG are shown in Table. \ref{tab:setting}. The final control input is $\V{u}\triangleq[a=u_0, \beta=(u_1-u_2)]^\intercal$, which is mapped to proper value in case the speed limit is exceeded.

\begin{figure}[h]
  \centering
  \includegraphics[width=3.4 in]{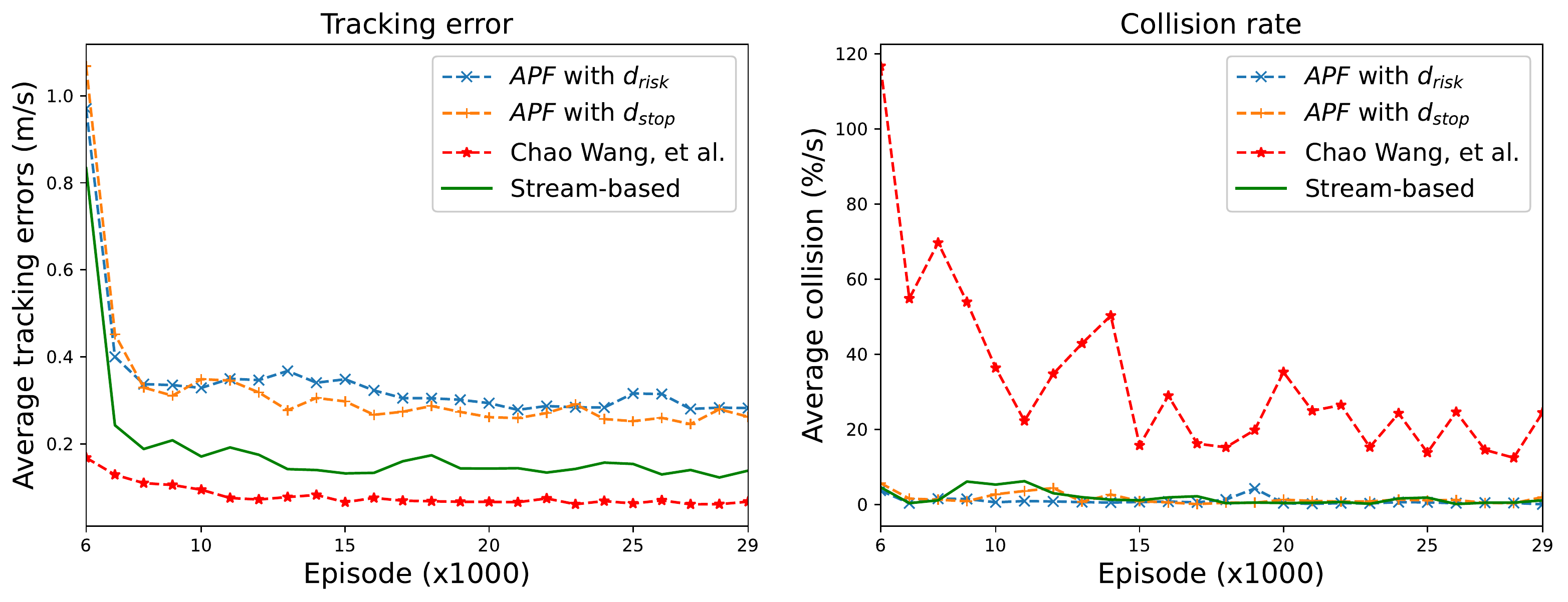}\\
  \centering
  \vspace{-4mm}
  \caption{Training curves of formation tracking with different avoidance cost, ignoring the warmup stages.}\label{fig:train_curve}
  \vspace{-4mm}
\end{figure}

Our method is compared with APF\cite{SamBavRamPueCam:C18} and the exponential reward proposed by Chao Wang, et al\cite{WanWanSheZha:J16}. The convergence curves of different methods are shown in Fig. \ref{fig:train_curve}. Since Chao Wang's reward is designed for large-scale obstacles, it fails to avoid scattered obstacles in our scene. APF in different safe distances and Stream-based method both ensure collision-free for more than $98\%$ during training, yet the later converges faster and achieves lower tracking error.
\begin{table}
\renewcommand{\arraystretch}{1.3}
\caption{Evaluation results}
\label{tab:result}
\begin{center}
\scalebox{0.7}{
\begin{tabular}{ccc}
\hline
Method&Tracking error $(m)$&Collision rate $(\%)$\\
\hline
APF, $d_{\text{risk}}$&$2.7704$&$6.46$\\
APF, $d_{\text{stop}}$&$1.9538$&$5.56$\\
Our method &$\textbf{1.3358}$&$\textbf{0.93}$\\
\hline
\end{tabular}}
\end{center}
\vspace{-8mm}
\end{table}

\begin{figure}[h]
  \centering
  \includegraphics[width=3.4 in]{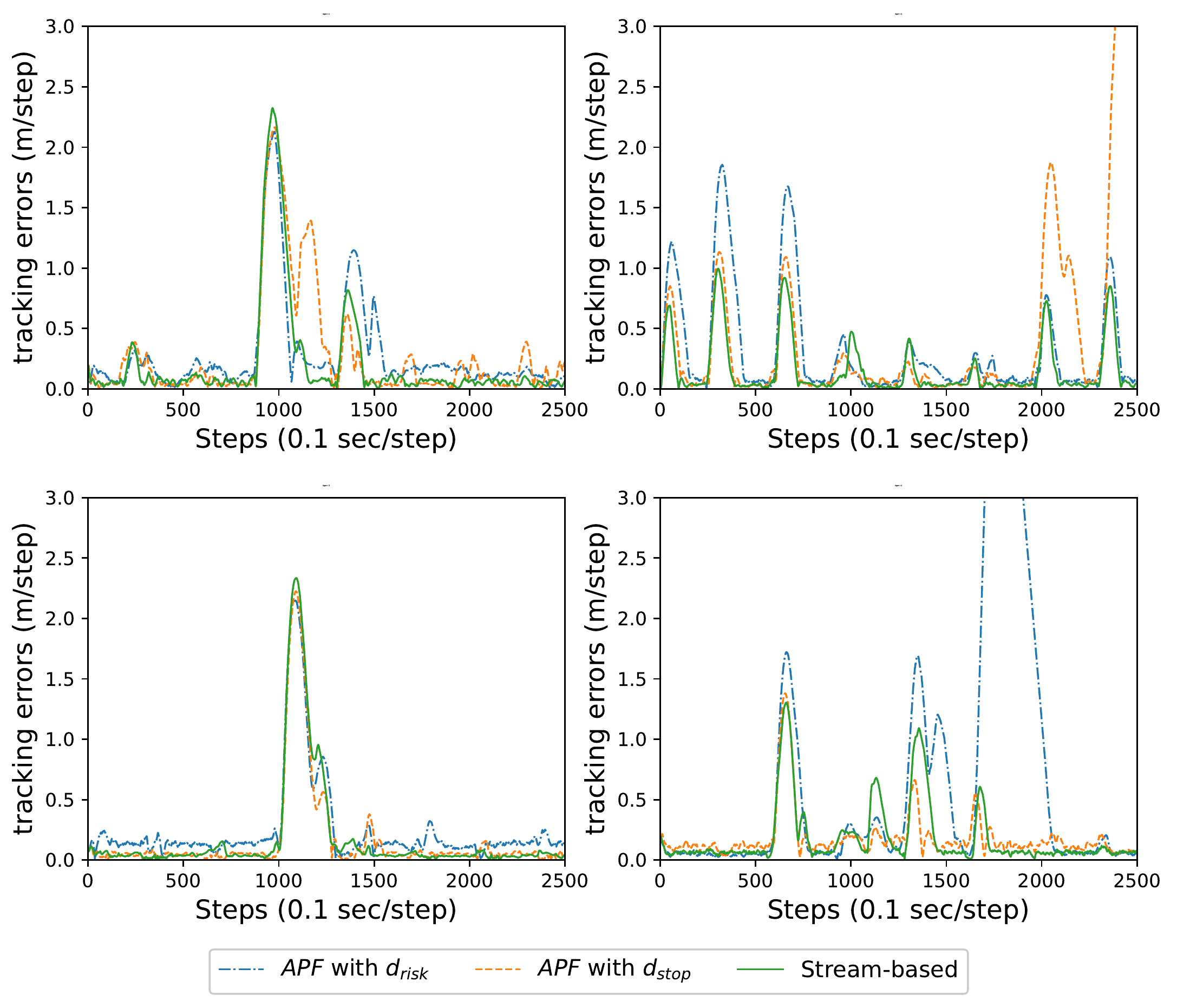}\\
  \centering
  \vspace{-4mm}
  \caption{The tracking errors under fixed environment.}\label{fig:eval_curve}
  \vspace{-4mm}
\end{figure}
To evaluate the robustness of our method, we increase the episode length to $2500$ steps and run $100$ times per method to obtain an average result. Table. \ref{tab:result} shows the average tracking error and collision rates of three models during evaluation. Fig. \ref{fig:eval_curve} depicts the entire tracking errors under the same obstacles deployment, in which the fluctuations are owing to avoidance. Through the elimination of unnecessary avoiding penalties, our method can return the correct formation position more efficiently without divergence. Thus, the robustness of our model when encountering multiple obstacles under long task duration is guaranteed. The entire simulation trajectories of the three models are shown in \url{https://youtu.be/jpsQ-kBJZk8}.
\vspace{-4mm}
\section{Conclusion and future work}
\vspace{-2mm}
In this paper, we propose a collision avoidance scheme for a formation navigation system based on the stream function. Unlike traditional stream-based methods, the proposed scheme avoids the requirement of global information by estimating the virtual flow field based on agents' local sensors. The numerical result reveals the improvement of our scheme in both collision rate and tracking error for $5\%$ and $0.6$m, respectively. In future work, we will investigate the potential of cooperation among the agents, along with the deployment of our algorithm on the real robot swarm system.
\vspace{-4mm}
\section*{Acknowledgment}
\vspace{-2mm}
This research was supported by the National Natural Science Foundation of China under Grant 61871256 Tsinghua University Initiative Scientific Research Program National Key R$\text{\&}$D Program of China 2020YFC1511803

\bibliographystyle{IEEEbib}
\bibliography{strings,refs,IEEEabrv,SGroupDefinition,SGroup}

\end{document}